%% file: icml2021.tex
\documentclass{article}

\usepackage{microtype}
\usepackage{graphicx}
\usepackage{subcaption}
\usepackage{booktabs}
\usepackage{hyperref}

\usepackage[usenames,dvipsnames]{xcolor}

\usepackage[accepted]{icml2021}

\usepackage{float}
\usepackage{soul}
\usepackage{titlecaps}
\usepackage{amsmath}
\usepackage{multirow, makecell}
\usepackage{amssymb}
\usepackage{pifont}
\newcommand{\circlemark}{\ding{108}}%

\definecolor{myred}{RGB}{227, 73, 75}
\definecolor{mygreen}{RGB}{34, 139, 34}

\newlength{\mysize}
\newcommand{\mycfs}[1]{\setlength{\mysize}{#1pt}%
  \fontsize{\mysize}{1.2\mysize}\selectfont}
  
\newif\ifcomments
\ifcomments
    \providecommand{\eric}[2][]{{\protect\color{red}{[Eric:\textbf{#1} #2]}}}
    \providecommand{\tony}[2][]{{\protect\color{orange}{[Tony:\textbf{#1} #2]}}}
    \providecommand{\sameer}[2][]{{\protect\color{violet}{[Sameer:\textbf{#1} #2]}}}
    \providecommand{\shi}[2][]{{\protect\color{violet}{[Shi:\textbf{#1} #2]}}}
    \providecommand{\dan}[2][]{{\protect\color{violet}{[Dan:\textbf{#1} #2]}}}
\else
    \providecommand{\eric}[2][]{}
    \providecommand{\tony}[2][]{}
    \providecommand{\sameer}[2][]{}
    \providecommand{\shi}[2][]{}
    \providecommand{\dan}[2][]{}
\fi

\newcommand{\methodname}{contextual calibration}
\newcommand{\methodnamebos}{Contextual calibration}

\newcommand{\mb}[1]{\boldsymbol{\mathbf{#1}}}

\icmltitlerunning{Calibrate Before Use: Improving Few-Shot Performance of Language Models}
\begin{document}
\twocolumn[
\icmltitle{Calibrate Before Use:\\Improving Few-Shot Performance of Language Models}

\icmlsetsymbol{equal}{*}
\begin{icmlauthorlist}
\icmlauthor{Tony Z. Zhao}{equal,berk}
\icmlauthor{Eric Wallace}{equal,berk}
\icmlauthor{Shi Feng}{umd}
\icmlauthor{Dan Klein}{berk}
\icmlauthor{Sameer Singh}{uci}
\end{icmlauthorlist}

\icmlaffiliation{berk}{UC Berkeley}
\icmlaffiliation{umd}{University of Maryland}
\icmlaffiliation{uci}{UC Irvine}

\icmlcorrespondingauthor{Eric Wallace}{\href{mailto:ericwallace@berkeley.edu}{ericwallace@berkeley.edu}}

\icmlkeywords{GPT-3, Few-shot learning, NLP, Transformers}
\vskip 0.3in
]
\printAffiliationsAndNotice{\icmlEqualContribution}

\begin{abstract}
GPT-3 can perform numerous tasks when provided a natural language prompt that contains a few training examples.
We show that this type of few-shot learning can be unstable: the choice of prompt format, training examples, and even the order of the training examples can cause accuracy to vary from near chance to near state-of-the-art. 
We demonstrate that this instability arises from the bias of language models towards predicting certain answers, e.g., those that are placed near the end of the prompt or are common in the pre-training data.
To mitigate this, we first estimate the model's bias towards each answer by asking for its prediction when given the training prompt and a content-free test input such as ``\texttt{N/A}''.
We then fit calibration parameters that cause the prediction for this input to be uniform across answers.
On a diverse set of tasks, this \emph{\methodname{}} procedure substantially improves GPT-3 and GPT-2's average accuracy (up to 30.0\% absolute) and reduces variance across different choices of the prompt.
\end{abstract}

\input{sections/10-intro}
\input{sections/20-setup}
\input{sections/30-variance}
\input{sections/40-analyzing}
\input{sections/50-normalization}
\input{sections/60-discussion}
\input{sections/70-related}
\input{sections/80-conclusion}
\input{sections/acknowledgements}
\bibliography{journal-abbrv,bib}
\bibliographystyle{icml2021}
\clearpage
\typeout{}

\appendix
\input{sections/99-appendix.tex}

\end{document}

%% file: sections/10-intro.tex
\begin{figure*}[t]
\captionsetup[subfigure]{labelformat=empty}
\begin{subfigure}{0.33\textwidth}
\centering
\includegraphics[width=1.0\textwidth]{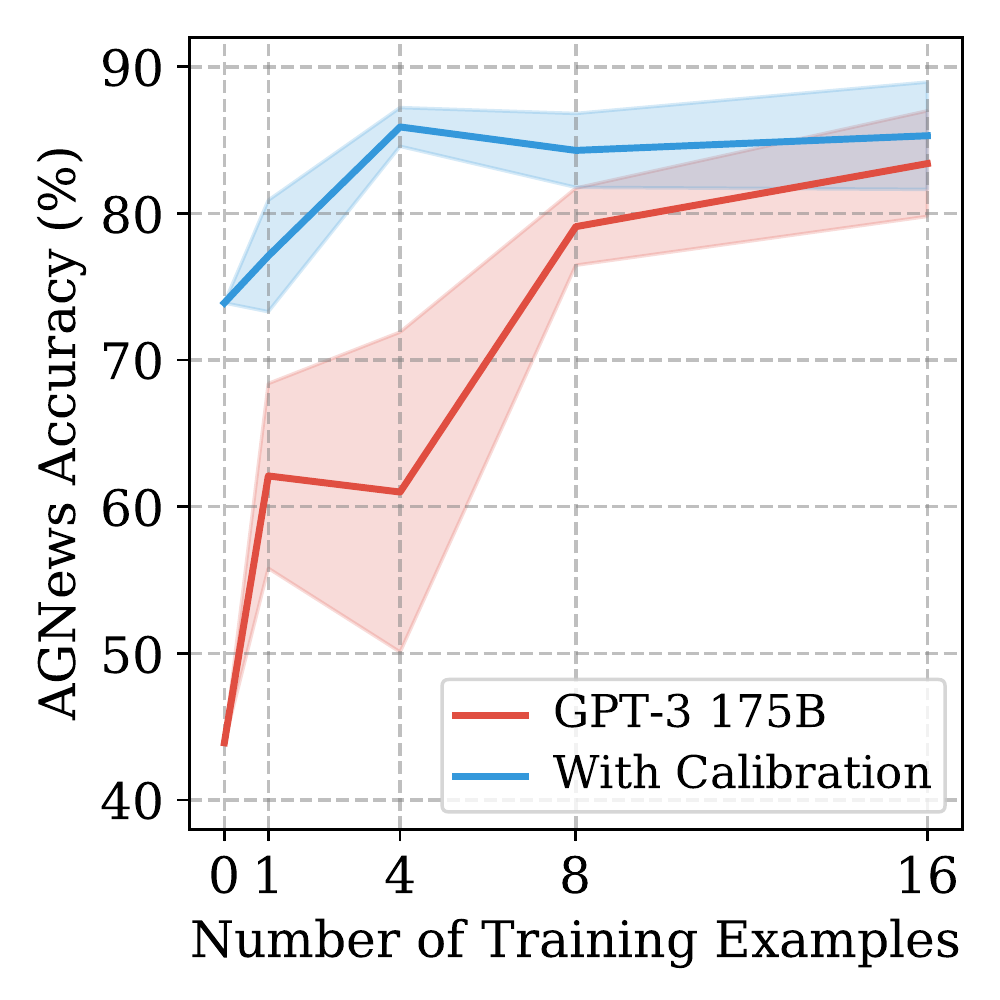}
\end{subfigure}%
\begin{subfigure}{0.33\textwidth}
\centering
\includegraphics[width=1.0\textwidth]{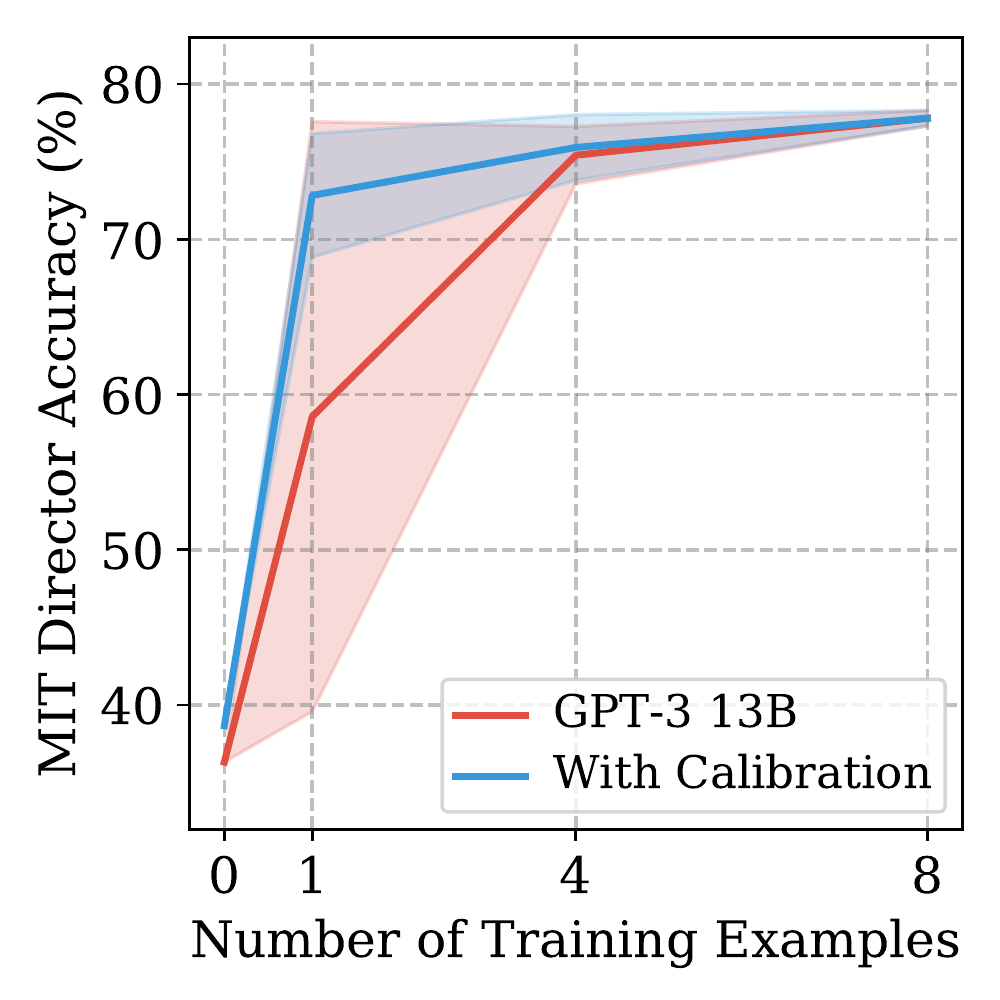}
\end{subfigure}%
\hfill
\begin{subfigure}{0.33\textwidth}
\centering
\includegraphics[width=1.0\textwidth]{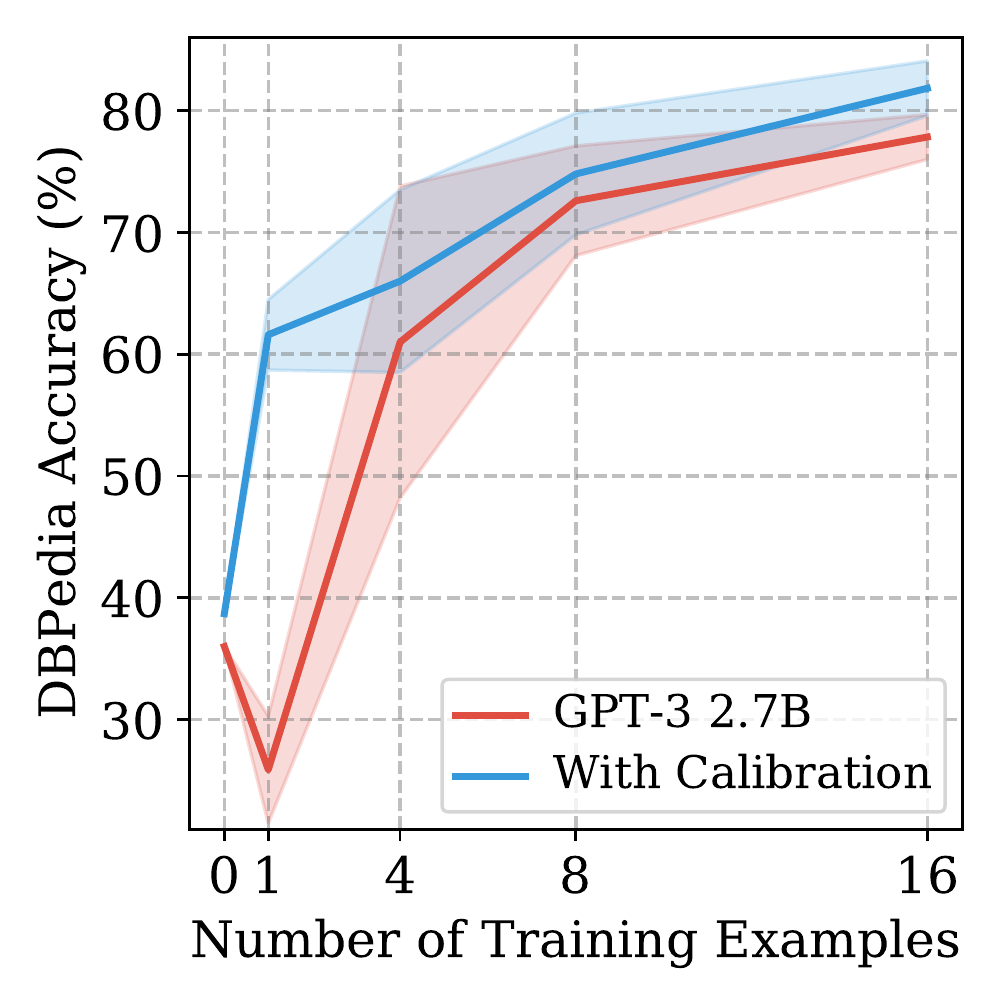}
\end{subfigure}%
\vspace{-0.33cm}
\caption{Few-shot learning can be highly unstable across different choices of the prompt. Above, we plot the mean accuracy ($\pm$ one standard deviation) across different choices of the training examples for three different datasets and model sizes. We show that our method, \emph{\methodname{}}, improves accuracy, reduces variance, and overall makes tools like GPT-3 more effective for end users.}
\label{fig:teaser}
\end{figure*}

\section{Introduction}\label{sec:intro}
Few-shot learning---the ability to learn tasks with limited examples---is an important aspect of intelligence~\citep{lake2015human,yogatama2019learning}.
Recent work shows that large neural language models can perform few-shot learning without finetuning~\citep{radford2019gpt2,brown2020language}.
Specifically, GPT-3~\citep{brown2020language} can perform numerous tasks when provided a few examples in a natural language \textit{prompt}. For example, to perform sentiment analysis one can condition GPT-3 on a prompt such as:

{
\begin{center}
\mycfs{10.0}
\vspace{-0.25cm}
{\usefont{T1}{cmss}{m}{n} \hspace{0.30cm}Input: Subpar acting. Sentiment: Negative}\newline

\vspace{-0.6cm}
{\usefont{T1}{cmss}{m}{n} \hspace{-0.52cm} Input: Beautiful film. Sentiment: Positive}

\vspace{-0.19cm}
{\usefont{T1}{cmss}{m}{n}\hspace{-1.74cm}Input: Amazing. \hspace{0.73cm}Sentiment:}
\end{center}
}
\vspace{-0.30cm}

where the first two lines correspond to two training examples and the last line is a test example. To make predictions, the model predicts whether the subsequent token is more likely to be the word ``Positive'' or ``Negative''.

This style of few-shot ``in-context'' learning is interesting because it shows that the model can learn without parameter updates. And, more importantly, it has numerous practical advantages over the now-standard approach of finetuning~\citep{radford2018improving,devlin2018BERT}. First, it allows practitioners to ``rapidly prototype'' NLP models: changing the prompt \textit{immediately} leads to a new model. Second, it provides a fully natural language interface to a machine learning model, which allows users---even those without technical expertise---to create NLP systems. 
Finally, since in-context learning reuses the same model for each task, it reduces memory requirements and system complexity when serving many different tasks.

However, despite these promises, we show that GPT-3's accuracy can be highly unstable across different prompts (Section~\ref{sec:variance}). A prompt contains three components: a format, a set of training examples, and a permutation (ordering) for those examples. We show that different choices for these factors can lead to highly different accuracies, e.g., changing the permutation of the training examples in a sentiment analysis prompt can change accuracy from near chance (54\%) to near state-of-the-art (93\%). This instability implies that GPT-3 users, who typically design prompts manually, cannot expect to consistently obtain good accuracy.

We next analyze what causes this instability. We identify three pitfalls of language models that lead them to be biased toward certain answers during few-shot learning. In particular, they suffer from majority label bias, recency bias, and common token bias (Section~\ref{sec:analysis}). The majority label and recency biases lead the model to predict training answers that appear frequently or near the end of the prompt. For example, a prompt that ends with a Negative training example may cause a bias towards the Negative class. On the other hand, the common token bias leads the model to prefer answers that are frequent in its pre-training data, e.g., it prefers ``United States'' over ``Saint Lucia'', which is likely suboptimal for the task of interest.

We identify that these biases typically result in a shift in the output distribution of the model. We can thus counteract these biases by ``calibrating'' the output distribution. Concretely, we estimate the model's bias towards certain answers by feeding in a dummy test input that is \textit{content-free}. 
In the prompt above for example, if we replace ``Amazing.'' with the string ``N/A'', the model predicts 62\% Positive. We then fit the calibration parameters so that the content-free input has uniform scores for each answer.
This \emph{\methodname{}} procedure provides a good setting of the calibration parameters without additional training data.

We test the effectiveness of \methodname{} on a range of tasks (Section~\ref{sec:calibration}). \methodnamebos{} consistently improves GPT-3 and GPT-2's accuracy (up to 30.0\% absolute) across different choices of the prompt format and examples (e.g., Figure~\ref{fig:teaser}). It also makes the accuracy more stable across different prompts, thus mitigating the need for prompt engineering. Overall, \methodname{} is a simple method that makes language models better few-shot learners: it enables end users to obtain higher accuracy with considerably less effort.

%% file: sections/20-setup.tex
\section{Background and Experimental Setup}

Neural autoregressive language models (LMs) take as input a sequence of tokens and output a probability distribution over the next token. Large neural LMs can perform tasks in a zero- or few-shot manner using in-context learning~\citep{radford2019gpt2,brown2020language}. To do so, a natural language \textit{prompt} is fed into the model. This prompt contains three components: a format, a set of training examples, and a permutation (ordering) of the training examples.

\noindent \textbf{Prompt Format} The prompt \textit{format} is a template which consists of placeholders for the training and test example(s) and possibly a natural language description of the task. For example, the format of the prompt in Section~\ref{sec:intro} is a template with the style: ``Input:'' \texttt{input} ``Sentiment:'' \texttt{label}. Many alternate formats exist, e.g., one could frame the task as question answering.

\noindent \textbf{Prompt Training Examples} The prompt's \textit{training examples} are used to teach the LM how to solve the task at hand. The prompt from Section~\ref{sec:intro} consists of two training examples; we refer to this as ``two-shot'' learning. We also consider ``zero-shot'' learning, where no training examples are present.

\noindent \textbf{Training Example Permutation} When training examples are used, they have a particular \textit{permutation}, e.g., the ``Subpar acting'' example comes first in the prompt from Section~\ref{sec:intro}. The permutation matters because neural language models update their hidden states in a left-to-right-fashion.

To make predictions on an input, we slot it into the test placeholder and generate from the LM. For example, see the ``Amazing.'' test example in the prompt from Section~\ref{sec:intro}. For generation tasks, we generate greedily from the LM until it produces a newline character. For classification tasks, the probability for each class is given by the probability assigned to its associated \textit{label name}, e.g., the words ``Negative'' and ``Positive'' for sentiment classification. 

\begin{figure*}[t]
\centering
\begin{minipage}{0.492\textwidth}
\centering
\includegraphics[width=1.0\columnwidth]{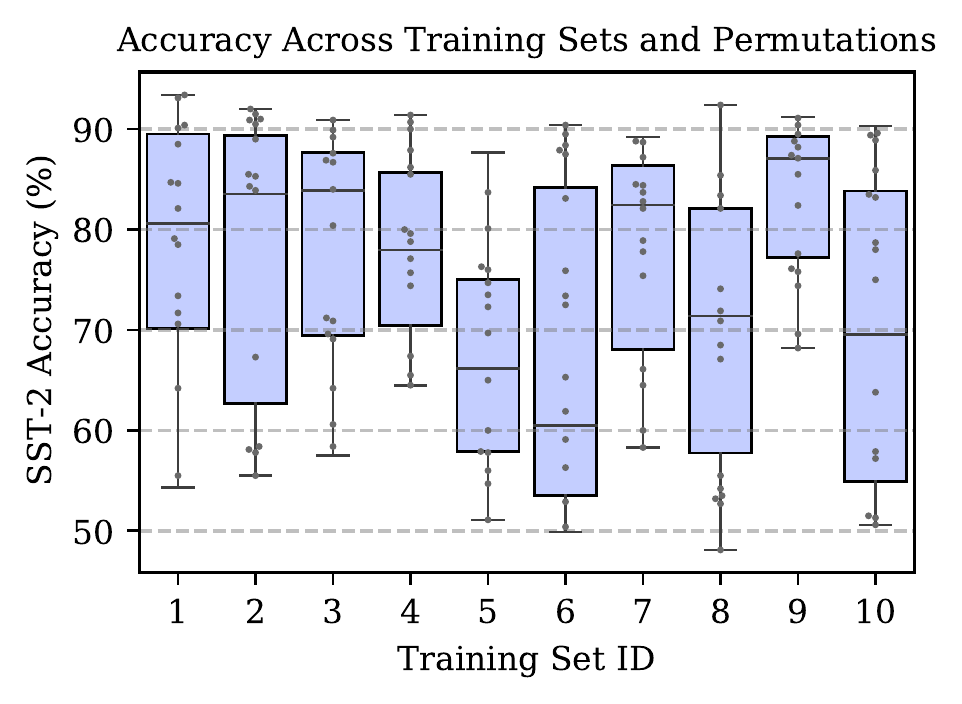}
\vspace{-0.99cm}
\caption{There is high variance in GPT-3's accuracy as we change the prompt's \textbf{training examples}, as well as the \textbf{permutation} of the examples. Here, we select ten different sets of four SST-2 training examples. For each set of examples, we vary their permutation and plot GPT-3 2.7B's accuracy for each permutation (and its quartiles).}
\label{fig:variance_training_set}
\end{minipage}
\hfill
\begin{minipage}{0.492\textwidth}
\centering
\vspace{-0.4cm}
\includegraphics[width=1.0\columnwidth]{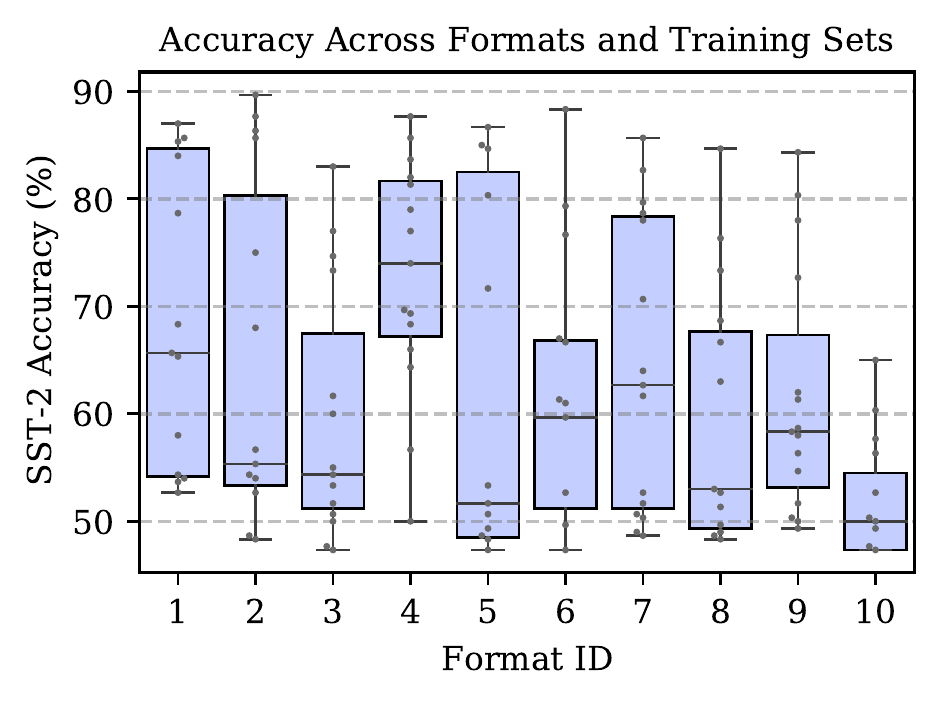}
\vspace{-0.99cm}
\caption{There is high variance in GPT-3's accuracy as we change the \textbf{prompt format}. In this figure, we use ten different prompt formats for SST-2. For each format, we plot GPT-3 2.7B's accuracy for different sets of four training examples, along with the quartiles.\eric{could add a title to theseto mimic the others}}
\label{fig:variance_format}
\end{minipage}
\end{figure*}

\subsection{Datasets and Prompt Formats}\label{subsec:datasets}

We use datasets for three tasks: text classification, fact retrieval, and information extraction. We use a fixed prompt format for each dataset unless otherwise specified. We show the format and examples from each dataset in Appendix~\ref{appendix:format}.

\noindent \textbf{Text Classification} We study text classification using six datasets: sentiment analysis using \textbf{SST-2}~\cite{socher2013recursive}, 6-way question classification using \textbf{TREC}~\cite{voorhees200trec}, textual entailment using 3-way \textbf{CB}~\cite{marneffe2019cb} and binary \textbf{RTE}~\cite{dagan2005pascal} from SuperGLUE~\cite{wang2019superglue}, and topic classification using the 4-way \textbf{AGNews}~\cite{zhang2015character} and 14-way \textbf{DBPedia}~\cite{zhang2015character} datasets. The prompt in Section~\ref{sec:intro} shows an example of the sentiment analysis task.

\noindent \textbf{Fact Retrieval} We evaluate fact retrieval with \textbf{LAMA} \cite{petroni2019language}. The dataset consists of knowledge base triples that are placed into templates with missing objects, e.g. ``Obama was born in''. We use these templates as our prompts, and remove the relations where the missing answer is not at the end of the template (left-to-right LMs cannot solve these). The answers are always single tokens, and we report average accuracy across all triples.

\noindent \textbf{Information Extraction} We consider information extraction using two slot filling datasets, \textbf{ATIS}~\citep{hemphill1990atis} and \textbf{MIT Movies} trivia10k13~\citep{liu2012conversational}. We use two random slots for each dataset, \textit{airline} and \textit{departure date} for ATIS, and \textit{director name} and \textit{movie genre} for MIT Movies. The answer for both datasets is a span of text from the input, e.g., the ATIS airline task is to predict ``american airlines'' when given the sentence ``list a flight on american airlines from toronto to san diego''. We use Exact Match between the model's generated output and the ground-truth span as our evaluation metric.

\subsection{Model Details}
We run our experiments on three sizes of GPT-3 (2.7B, 13B, and 175B parameters) as well as GPT-2 (1.5B parameters). We access GPT-3 using the OpenAI API. We release code to replicate our experiments.\footnote{\url{https://www.github.com/tonyzhaozh/few-shot-learning}}

%% file: sections/30-variance.tex
\section{Accuracy Varies Highly Across Prompts}\label{sec:variance}

\begin{figure*}[t]
\captionsetup[subfigure]{labelformat=empty}
\begin{subfigure}{1.0\textwidth}
\centering
\includegraphics[trim={0.0cm 0.68cm 0.0cm 0.0cm},clip,width=1.0\textwidth]{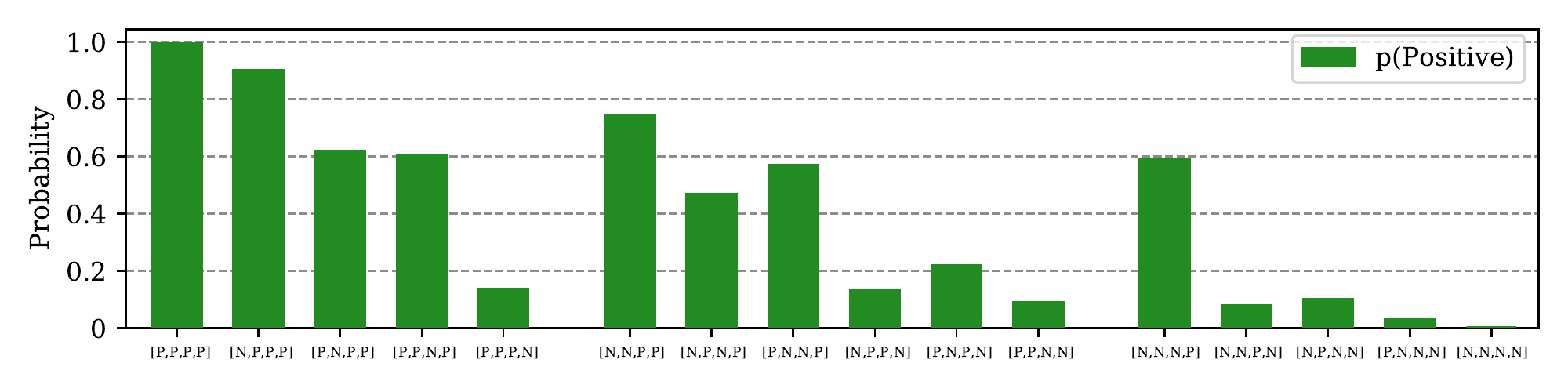}
\end{subfigure}%
\newline
\begin{subfigure}{1.0\textwidth}
\centering
\includegraphics[width=1.0\textwidth]{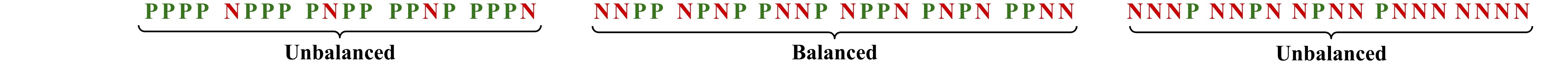}
\end{subfigure}%
\vspace{-0.20cm}
\caption{\textbf{Majority label and recency biases} cause GPT-3 to become biased towards certain answers and help to explain the high variance across different examples and orderings. Above, we use 4-shot SST-2 with prompts that have different class balances and permutations, e.g., [P P N N] indicates two positive training examples and then two negative. We plot how often GPT-3 2.7B predicts Positive on the balanced validation set. When the prompt is unbalanced, the predictions are unbalanced (\emph{majority label bias}). In addition, balanced prompts that have one class repeated near the end, e.g., end with two Negative examples, will have a bias towards that class (\emph{recency bias}).}
\label{fig:label_permutation_full}
\end{figure*}

This section studies how GPT-3's accuracy changes as we vary each aspect of the prompt (training examples, permutation, format). We focus on a subset of the datasets to simplify our analysis; in Section~\ref{sec:calibration} we show that our findings hold across all of the datasets we study.

\noindent \textbf{GPT-3's accuracy depends highly on both selection and permutation of training examples.}
Concretely, we use a fixed prompt format and choose different random sets of training examples. For each set of training examples, we evaluate the accuracy for all possible permutations.

Figure~\ref{fig:variance_training_set} shows the results for SST-2 (4-shot, GPT-3 2.7B). Surprisingly, varying the permutation can be as important, or even more important, than which training examples are chosen. For example, varying the permutation of the training examples can cause accuracy to go from near chance (54.3\%) to near state-of-the-art (93.4\%). For a qualitative example of the sensitivity to permutations, see Table~\ref{tab:qualitative} in Appendix~\ref{appendix:analyzing}. This high importance on example order is in contrast to standard machine learning, where the ordering of examples during training is typically an afterthought.

\noindent \textbf{The variance persists with more data and larger models.} Adding more training examples into the prompt does not necessarily reduce the variance in accuracy. We sweep over the number of training examples for three different datasets in Figure~\ref{fig:teaser} (red curves). The variance remains high even when we use 16 training examples. Moreover, adding more training examples can sometimes hurt accuracy (e.g., mean accuracy drops from 36.0\% to 25.9\% for DBPedia 0-shot to 1-shot).
The variance in accuracy can also remain high when using larger models, e.g., the left of Figure~\ref{fig:teaser}.

\noindent \textbf{GPT-3's accuracy depends highly on prompt format.} We next keep the set of training examples and permutations fixed but vary the prompt format. We focus on SST-2, and we manually design an additional 14 prompt formats. The formats include question-answer templates, conversation-style templates, prompts that resemble Web pages, and variations on the label names (all formats available in Table~\ref{tab:sst2_format_exploration} in Appendix~\ref{appendix:format}). The accuracy for ten of the formats is shown in Figure~\ref{fig:variance_format}. We find that some of the formats are better than others on average. However, all of the formats still suffer from high variance across different training sets.

%% file: sections/40-analyzing.tex
\section{What Causes the High Variance?}\label{sec:analysis}

We next analyze \emph{why} GPT-3's accuracy varies across different training examples, permutations, and prompt formats. Concretely, we show that the variance arises because LMs are biased towards outputting answers that are (1) frequent in the prompt (majority label bias), (2) towards the end of the prompt (recency bias), and (3) common in the pre-training data (common token bias).

\noindent \textbf{Majority Label Bias} We find that GPT-3 is biased towards answers that are frequent in the prompt. A trivial case is when a text classification prompt has a class imbalance, e.g., more Positive than Negative sentiment examples. This is demonstrated in the ``unbalanced'' region of Figure~\ref{fig:label_permutation_full}: when one class is more common, GPT-3 2.7B is heavily biased towards predicting that class. Since the SST-2 sentiment analysis dataset is balanced, this bias causes large accuracy degradations. The majority label bias also explains why we frequently observe a drop in accuracy when moving from 0-shot to 1-shot---we found that the drop is due to the model frequently repeating the class of the one training example.

The majority label bias also occurs for generation tasks. On the validation set for 4-shot LAMA with GPT-3 2.7B, 50.2\% of the model predictions are a repeat of one of the four training answers (the correct repeat rate is 24.7\%). Overall, the majority label bias helps to explain why different choices for the training examples heavily influence GPT-3's accuracy---it shifts the distribution of model predictions.

\noindent \textbf{Recency Bias} The model's majority label bias is aggravated by its \emph{recency bias}: the tendency to repeat answers that appear towards the end of the prompt. The ``balanced'' region of Figure~\ref{fig:label_permutation_full} demonstrates this. For instance, when two Negative examples appear at the end (P P N N), the model will heavily prefer the Negative class. Moreover, the recency bias can outweigh the majority label bias, e.g., the ``P P P N'' training set leads to nearly 90\% of predictions being Negative, despite $\frac{3}{4}$ of the training examples being Positive.

Recency bias also affects generation tasks. For 4-shot LAMA, the training answers that are closer to the end of the prompt are more likely to be repeated by the model. Concretely, the model ``overpredicts'' the answer from the 1st, 2nd, 3rd, and 4th training example by 8.5\%, 8.3\%, 14.3\%, and 16.1\%, respectively.\footnote{Over all relations, as well as three different sets of training examples, the model repeats the training example at a rate of 20.7\%, 19.8\%, 29.9\%, and 26.8\%. The ground-truth repeat rate is 12.2\%, 11.5\%, 15.6\%, and 10.7\%. We define ``overpredicts'' as the model's repeat rate minus the ground-truth repeat rate.} Overall, recency bias helps to explain why the \textit{permutation} of the training examples is important---the ordering  of the examples heavily influences the distribution of the model predictions.

\noindent \textbf{Common Token Bias} Finally, we find that GPT-3 is biased towards outputting tokens that are common in its \textit{pre-training} distribution, which is likely suboptimal for the distribution of answers on the \textit{downstream} task.
A simple case of this occurs for the LAMA fact retrieval dataset, where the model often predicts common entities such as ``America'' when the ground-truth answer is instead a rare entity. 

A more nuanced case of the common token bias occurs for text classification. Recall that the model makes predictions by generating the label name associated with each class. Because certain label names appear more frequently in the pre-training data, the model will be inherently biased towards predicting certain classes. For example, on DBPedia (a balanced 14-way topic classification dataset), GPT-3 predicts the ``book'' class 11$\times$ more often than the ``artist'' class. In fact, there is a moderate correlation ($r=0.67$) between the frequency of a DBPedia label name and the rate at which GPT-3 predicts its class.\footnote{The frequency of a token on the web is calculated using Google Ngrams \url{https://books.google.com/ngrams}. The predictions are from the 0-shot setting on the validation set.}
Overall, the common token bias helps to explain why the choice of label names is important, and why the model struggles on rare answers.

\begin{figure}[!h]
\vspace{-0.10cm}
\centering
\includegraphics[trim={0.0cm 0.0cm 0.0cm 0.65cm},clip,width=1.0\columnwidth]{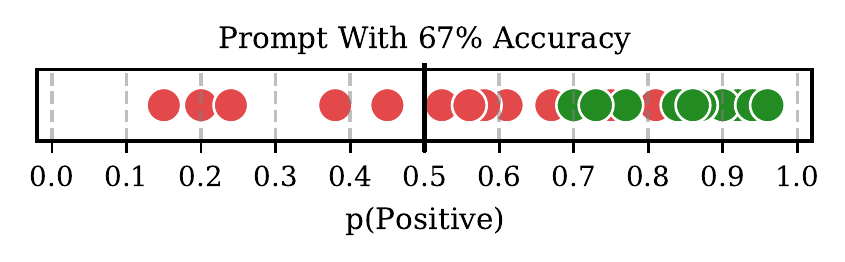}
\vspace{-1.0cm}
\caption{The Positive class probability for 25 random test inputs for a particular sentiment analysis prompt. Negative ground-truth examples are marked with \textbf{\textcolor{myred}{\circlemark}} and Positive are marked with \textbf{\textcolor{mygreen}{\circlemark}}.}
\vspace{-0.15cm}
\label{fig:decision_boundary}
\end{figure}

\noindent \textbf{The Impact of Biases on Model Predictions} We find that the end result of the above three biases is typically a simple shift in the model's output distribution. For example, Figure~\ref{fig:decision_boundary} visualizes this shift for a SST-2 sentiment prompt.

The prompt used in Figure~\ref{fig:decision_boundary} and the model's intrinsic biases cause it to frequently predict high confidence for the Positive class. Since the default 50\% threshold is used to make predictions, this results in frequent false positives. Importantly, note that if we could optimally set the classification threshold (p(Positive) = 0.68 in this case), the classifier would be highly accurate (94\% on the validation set).

%% file: sections/50-normalization.tex
\section{ \titlecap{\methodname{}}}\label{sec:calibration}

Thus far, we have shown that GPT-3 is biased towards certain answers due to the prompt and the model's intrinsic biases. Here, we look to correct this by ``calibrating'' the model's output probabilities.\footnote{The output of GPT-3 is biased (its outputs are shifted), similar to how measurement devices such as voltage meters or weighing scales are biased. Just like how these devices require ``calibration before use'', where the devices' outputs are scaled/zeroed-out, we hope to apply a similar calibration procedure to LMs. This goal is distinct from statistical calibration~\cite{brier1950verification,guo2017calibration}, i.e., aligning a model's confidence estimate with its true accuracy.}
A common technique for adjusting output probabilities is to apply an affine transformation~\cite{platt1999scaling,guo2017calibration}:
\setlength{\abovedisplayskip}{4pt}
\setlength{\belowdisplayskip}{4pt}
\begin{equation}
\mb{\hat{q}} = \text{softmax}(\mb{W}\mb{\hat{p}} + \mb{b}),
\end{equation}
where a weight matrix $\mb{W}$ and a bias vector $\mb{b}$ are applied to the original probabilities $\mb{\hat{p}}$ to get the new probabilities $\mb{\hat{q}}$.\footnote{This affine transformation is usually applied to the logits, i.e., prior to the softmax. However, we only have access to GPT-3's output probabilities in the OpenAI API.} For classification tasks, $\mb{\hat{p}}$ is the set of probabilities that are associated with each label name, renormalized to one. For generation tasks, $\mb{\hat{p}}$ is the entire set of probabilities for the first token.\footnote{We only calibrate the prediction of the first output token for generation tasks. This is reasonable because, for the tasks we consider, we found that the model's predictions are highly deterministic after generating the first token.} In this paper, we restrict the matrix $\mb{W}$ to be diagonal, known as vector scaling~\cite{guo2017calibration}, to prevent the parameters from growing quadratically in the size of $\mb{\hat{p}}$ (which is $\approx$ $50,000$ for generation tasks). 

The main challenge in the zero- or few-shot setting is that we do not have data to learn $\mb{W}$ and $\mb{b}$. We thus propose a novel data-free procedure to infer a good setting of these parameters. The key idea is that the model's bias towards certain answers can be estimated by feeding in a \textit{content-free} input such as the string ``N/A''. For example, consider the two-shot prompt:
{
\begin{center}
\mycfs{10.0}
\vspace{-0.25cm}
{\usefont{T1}{cmss}{m}{n} \hspace{0.30cm}Input: Subpar acting. Sentiment: Negative}\newline

\vspace{-0.6cm}
{\usefont{T1}{cmss}{m}{n} \hspace{-0.52cm} Input: Beautiful film. Sentiment: Positive}

\vspace{-0.19cm}
{\usefont{T1}{cmss}{m}{n}\hspace{-1.72cm}Input: N/A \hspace{1.5cm}Sentiment:}
\end{center}
}
\vspace{-0.21cm}
where ``N/A'' serves as the test input. Ideally, GPT-3 would score this test input as 50\% Positive and 50\% Negative. However, the model's biases cause it to score this input as 61.8\% Positive. Note that this error is \textit{contextual}: a different choice of the training examples, permutation, and format will lead to different predictions for the content-free input.

We can correct this error by setting $\mb{W}$ and $\mb{b}$ so that the class scores for the content-free input are uniform. We first obtain $\mb{\hat{p}}$ for the content-free input, denoted $\mb{\hat{p}}_{\text{cf}}$.
We then set $\mb{W} = \text{diag}(\mb{\hat{p}}_{\text{cf}})^{-1}$ 
and $\mb{b}$ to the all-zero vector.\footnote{An alternate solution is to set $\mb{b}$ to $-\mb{\hat{p}}_{\text{cf}}$ and $\mb{W}$ to the identity. Empirically, this alternate solution yields higher accuracy for generation tasks (where the dimensionality of $\mb{\hat{p}}$ is large). The solution in the main text performs better for classification.} To make test predictions, we compute $\mb{W}\mb{\hat{p}} + \mb{b}$ and take the argmax. 

\noindent \textbf{Implementation Details} This \textit{\methodname{}} procedure adds trivial amounts of computational overhead and is implemented in a few lines of code (compute and save $\mb{\hat{p}}_{\text{cf}}$, adjust output probabilities). For the content-free input, many good choices exist, including ``N/A'', the empty string, and gibberish tokens. In all our experiments, we average the probabilities from three content-free inputs: ``N/A'', ``[MASK]'', and the empty string.\footnote{We found this simple ensemble to achieve the best results for AGNews, and we reuse it for all other datasets. See Section~\ref{subsec:ablations} for an ablation on the choice of content-free input.} One could also craft the content-free input in a task-specific manner. We explore this for LAMA, where we replace the subject with the content-free input, e.g., we use ``N/A was born in'' as the input.

\input{sections/52-table}

\subsection{Results for Contextual Calibration}\label{subsec:results}

Here, we evaluate the effectiveness of \methodname{} across all of our datasets and LMs. We first use a fixed prompt format and select five different random sets of training examples, placing them in an arbitrary order in the prompt. We do not artificially balance the labels of the training examples for the classification tasks. We use the same sets of training examples for the baseline (standard decoding without calibration) and \methodname{}. We use labeling budgets of 0--8 examples; using more than 8-shots causes the cost of querying the OpenAI API to become prohibitively expensive.

Table~\ref{table:main_results} shows the results and Figure~\ref{fig:teaser} in Section~\ref{sec:intro} plots the same data for a subset of the tasks.

\noindent \textbf{Improves Mean And Worst-Case Accuracy} \methodnamebos{} dramatically improves GPT-3's average and worst-case accuracy, by up to 30.0\% absolute. These gains hold for both classification and generation tasks. \methodnamebos{} also sometimes allows GPT-3 2.7B to outperform the GPT-3 175B baseline---by up to 19.3\%---despite being over 50x smaller. 

\begin{figure}[t]
\centering
\includegraphics[trim={0.0cm 0cm 0.0cm 0cm},clip, width=0.99\linewidth]{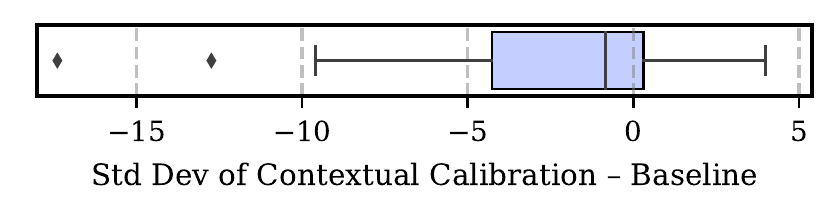}
\vspace{-0.55cm}
\caption{Aside from improving mean accuracy, \methodname{} also reduces the standard deviation of accuracy across different choices of the training examples. We plot the difference in standard deviation between \methodname{} and the baseline from Table~\ref{table:main_results}.}
\label{fig:variance_reduction}
\end{figure}

\noindent \textbf{Can Reduce Variance Across Training Sets} Figure~\ref{fig:variance_reduction} plots the difference in the standard deviation between the baseline and \methodname{} for all tasks from Table~\ref{table:main_results}. \methodnamebos{} reduces the variance considerably in a majority of cases, and it does not increase variance by much in the remaining cases.

\noindent \textbf{Reduces Drop from 0-shot to 1-shot} For the baseline, there are four cases where there is a drop in accuracy when moving from 0-shot to 1-shot (TREC, AGNews, DBpedia, SST-2). We attribute this drop to the majority label bias (see discussion in Section~\ref{sec:analysis}). Calibration removes this drop in three out of four cases.

\noindent \textbf{Improves GPT-2} We also test GPT-2 1.5B (see  Table~\ref{table:main_results_gpt2} in Appendix~\ref{appendix:analyzing}). We find that like GPT-3, GPT-2's accuracy also highly varies across different prompts. This suggests that the variance that we observe for few-shot in-context learning is a general problem for LMs. Second, \methodname{} works out-of-the-box for GPT-2---it improves the mean accuracy and reduces variance for most tasks.

\begin{figure}[t]
\centering
\includegraphics[trim={0.4cm 1.2cm 0.3cm 1.25cm},clip, width=1.0\linewidth]{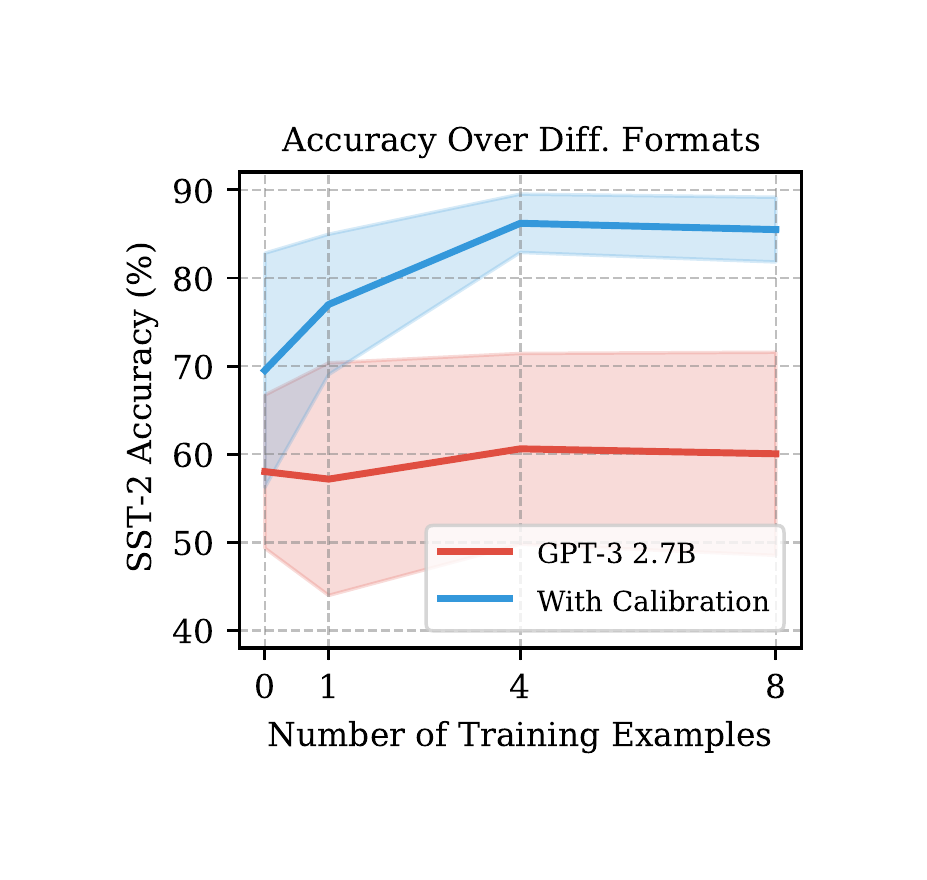}
\vspace{-0.80cm}
\caption{GPT-3 has high variance across different prompt formats; \methodname{} reduces this variance and improves mean accuracy. We show the mean accuracy ($\pm$ standard deviation) over 15 different prompt formats for SST-2.}
\label{fig:format_after_calibration}
\end{figure}

\noindent \textbf{Improves Accuracy Across Formats} In our next set of experiments, we use a fixed set of training examples and vary the prompt format. We use the 15 prompt formats for SST-2 discussed in Section~\ref{sec:variance}. We also create 15 prompt formats for each of three random relations in LAMA (P20, P159, P19) by using the paraphrases of the original LAMA templates generated by \citet{jiang2019can}. 
Figure~\ref{fig:format_after_calibration} shows the results before and after calibration for SST-2, and Figure~\ref{fig:format_appendix} in Appendix~\ref{appendix:analyzing} show the results for LAMA.
\methodnamebos{} improves the average and worst-case accuracy for both datasets, and reduces the variance for SST-2. 

\subsection{Ablations on \titlecap{\methodname{}}}\label{subsec:ablations}

\begin{figure}
\centering
\includegraphics[trim={0.4cm 1.2cm 0.2cm 1.25cm},clip, width=1.0\linewidth]{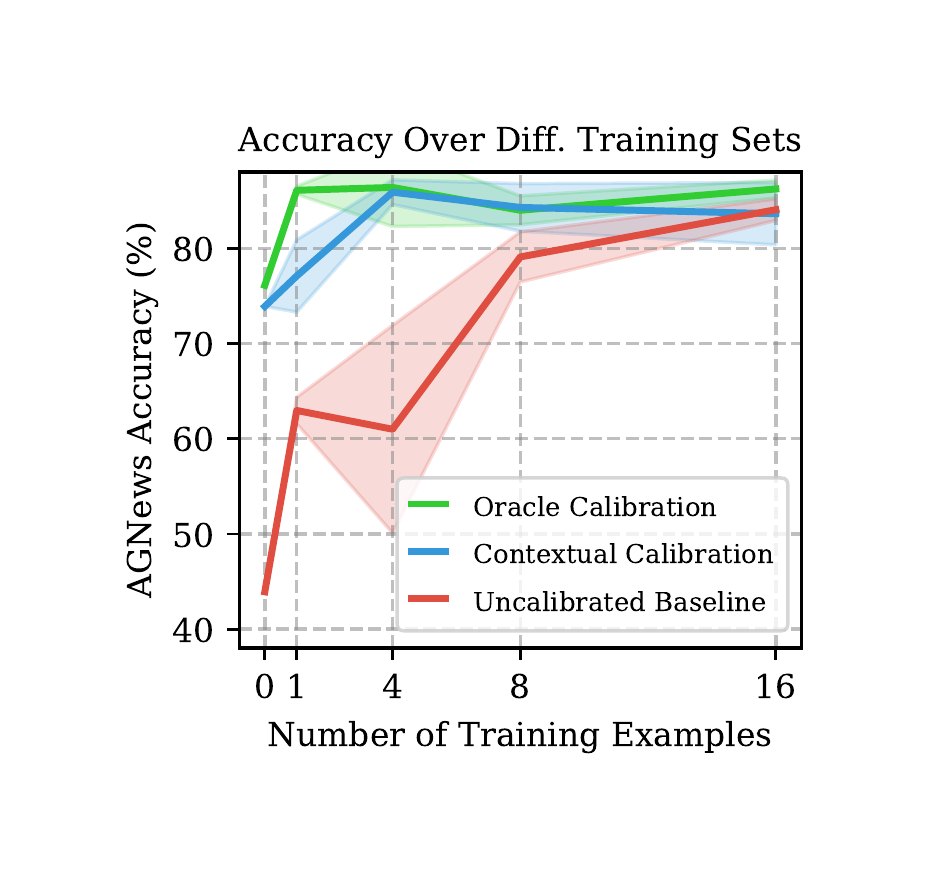}
\vspace{-0.81cm}
\caption{\methodnamebos{}, despite using no training data, achieves similar accuracy to an ``oracle'' calibration that finds the best $\mb{W}$ using the validation set. The plot shows GPT-3 175B's mean accuracy ($\pm$ standard deviation)  on AGNews over different choices of the training examples.}
\label{fig:oracle_versus_us}
\end{figure}

We finally conduct two analyses/ablations on \methodname{}.  We first analyze how effective \methodname{} is at inferring a good setting of $\mb{W}$. To do so, we compare its accuracy to an ``oracle calibration'' method that uses the validation set to find the best possible diagonal $\mb{W}$. We evaluate this oracle on AGNews, and find that \methodname{} is surprisingly close to it (Figure~\ref{fig:oracle_versus_us}).

We also study how the choice of content-free input affects accuracy. In Table~\ref{table:context_words} in Appendix~\ref{appendix:analyzing}, we show the accuracy for SST-2 and AGNews for different choices of the content-free input. The choice of content-free input matters, however, many good choices exist.

%% file: sections/52-table.tex
\begin{table*}[t]
\centering
\mycfs{10}
\setlength{\tabcolsep}{5pt}
\begin{tabular}{lcllllllll}
\toprule
\textbf{Dataset} & \textbf{LM} &  \multicolumn{2}{c}{\textbf{0-shot}} &       \multicolumn{2}{c}{\textbf{1-shot}} & \multicolumn{2}{c}{\textbf{4-shot}} & \multicolumn{2}{c}{\textbf{8-shot}} \\ 
\cmidrule(lr){3-4}
\cmidrule(lr){5-6}
\cmidrule(lr){7-8}
\cmidrule(lr){9-10}
& & Baseline & {~~Ours} & Baseline & {~~~Ours} & Baseline & {~~~Ours} & Baseline & {~~~Ours} \\
\midrule
\multicolumn{5}{l}{\hspace{-0.22cm} \mycfs{10} \emph{Text Classification}} \\
\multirow{2}{*}{AGNews} & $2.7$B & $44.7_{\hspace{0.05cm}0.0}$ & $\textbf{63.2}_{\hspace{0.05cm}0.0}$ & $33.0_{\hspace{0.05cm}5.1}$ & $\textbf{59.6}_{\hspace{0.05cm}6.4}$ & $43.3_{\hspace{0.05cm}8.3}$ & $\textbf{71.1}_{\hspace{0.05cm}8.5}$ & $50.8_{\hspace{0.05cm}7.8}$ & $\textbf{72.7}_{\hspace{0.05cm}5.8}$ \\
\vspace{0.09cm}
& $175$B & $43.9_{\hspace{0.05cm}0.0}$ & $\textbf{73.9}_{\hspace{0.05cm}0.0}$ & $62.1_{\hspace{0.05cm}6.3}$ & $\textbf{77.1}_{\hspace{0.05cm}3.8}$ & $61.0_{\hspace{0.05cm}10.9}$ & $\textbf{85.9}_{\hspace{0.05cm}1.3}$ & $79.1_{\hspace{0.05cm}2.6}$ & $\textbf{84.3}_{\hspace{0.05cm}2.5}$ \\
\multirow{2}{*}{TREC} & $2.7$B & $31.0_{\hspace{0.05cm}0.0}$ & $\textbf{38.8}_{\hspace{0.05cm}0.0}$ & $24.3_{\hspace{0.05cm}6.4}$ & $\textbf{36.8}_{\hspace{0.05cm}7.7}$ & $25.8_{\hspace{0.05cm}11.5}$ & $\textbf{38.6}_{\hspace{0.05cm}13.2}$ & $29.3_{\hspace{0.05cm}8.0}$ & $\textbf{44.3}_{\hspace{0.05cm}11.4}$ \\
\vspace{0.09cm}
& $175$B & $47.4_{\hspace{0.05cm}0.0}$ & $\textbf{57.4}_{\hspace{0.05cm}0.0}$ & $57.7_{\hspace{0.05cm}6.0}$ & $\textbf{75.7}_{\hspace{0.05cm}1.4}$ & $60.2_{\hspace{0.05cm}7.6}$ & $\textbf{69.7}_{\hspace{0.05cm}1.4}$ & $45.6_{\hspace{0.05cm}4.0}$ & $\textbf{66.9}_{\hspace{0.05cm}6.5}$ \\
\multirow{2}{*}{CB} & $2.7$B & $44.6_{\hspace{0.05cm}0.0}$ & $\textbf{50.0}_{\hspace{0.05cm}0.0}$ & $\textbf{33.8}_{\hspace{0.05cm}16.6}$ & $33.0_{\hspace{0.05cm}7.3}$ & $43.5_{\hspace{0.05cm}11.9}$ & $\textbf{54.2}_{\hspace{0.05cm}4.7}$ & $43.9_{\hspace{0.05cm}8.4}$ & $\textbf{53.0}_{\hspace{0.05cm}7.7}$ \\
\vspace{0.09cm}
& $175$B & $30.4_{\hspace{0.05cm}0.0}$ & $\textbf{48.2}_{\hspace{0.05cm}0.0}$ & $50.9_{\hspace{0.05cm}6.7}$ & $\textbf{51.8}_{\hspace{0.05cm}7.2}$ & $45.2_{\hspace{0.05cm}19.4}$ & $\textbf{60.7}_{\hspace{0.05cm}6.7}$ & $59.6_{\hspace{0.05cm}11.3}$ & $\textbf{65.0}_{\hspace{0.05cm}7.9}$ \\
\multirow{2}{*}{RTE} & $2.7$B & $44.8_{\hspace{0.05cm}0.0}$ & $\textbf{49.5}_{\hspace{0.05cm}0.0}$ & $49.6_{\hspace{0.05cm}2.9}$ & $\textbf{50.4}_{\hspace{0.05cm}2.7}$ & $44.0_{\hspace{0.05cm}1.4}$ & $\textbf{54.5}_{\hspace{0.05cm}4.7}$ & $49.2_{\hspace{0.05cm}1.9}$ & $\textbf{54.8}_{\hspace{0.05cm}2.8}$ \\
& $175$B & $\textbf{57.8}_{\hspace{0.05cm}0.0}$ & $\textbf{57.8}_{\hspace{0.05cm}0.0}$ & $\textbf{62.9}_{\hspace{0.05cm}2.7}$ & $62.8_{\hspace{0.05cm}2.3}$ & $58.7_{\hspace{0.05cm}11.9}$ & $\textbf{60.4}_{\hspace{0.05cm}8.1}$ & $\textbf{66.2}_{\hspace{0.05cm}5.8}$ & $65.5_{\hspace{0.05cm}2.5}$ \\
\multirow{2}{*}{SST-2} & $2.7$B & $57.2_{\hspace{0.05cm}0.0}$ & $\textbf{71.4}_{\hspace{0.05cm}0.0}$ & $67.3_{\hspace{0.05cm}7.9}$ & $\textbf{79.1}_{\hspace{0.05cm}8.3}$ & $59.1_{\hspace{0.05cm}10.2}$ & $\textbf{79.9}_{\hspace{0.05cm}7.8}$ & $54.0_{\hspace{0.05cm}4.3}$ & $\textbf{82.0}_{\hspace{0.05cm}5.5}$ \\
\vspace{0.09cm}
& $175$B & $71.6_{\hspace{0.05cm}0.0}$ & $\textbf{75.8}_{\hspace{0.05cm}0.0}$ & $93.3_{\hspace{0.05cm}2.8}$ & $\textbf{94.7}_{\hspace{0.05cm}1.4}$ & $93.6_{\hspace{0.05cm}3.3}$ & $\textbf{94.3}_{\hspace{0.05cm}1.0}$ & $\textbf{95.6}_{\hspace{0.05cm}1.0}$ & $95.3_{\hspace{0.05cm}0.7}$ \\
\multirow{2}{*}{DBPedia} & $2.7$B & $36.0_{\hspace{0.05cm}0.0}$ & $\textbf{38.7}_{\hspace{0.05cm}0.0}$ & $25.9_{\hspace{0.05cm}4.4}$ & $\textbf{61.6}_{\hspace{0.05cm}2.9}$ & $61.0_{\hspace{0.05cm}12.8}$ & $\textbf{66.0}_{\hspace{0.05cm}7.5}$ & $72.6_{\hspace{0.05cm}4.5}$ & $\textbf{74.8}_{\hspace{0.05cm}5.0}$ \\
\vspace{0.09cm}
& $175$B & $22.0_{\hspace{0.05cm}0.0}$ & $\textbf{59.7}_{\hspace{0.05cm}0.0}$ & $79.3_{\hspace{0.05cm}3.0}$ & $\textbf{85.3}_{\hspace{0.05cm}2.2}$ & $84.6_{\hspace{0.05cm}5.8}$ & $\textbf{86.9}_{\hspace{0.05cm}4.0}$ & $82.3_{\hspace{0.05cm}7.8}$ & $\textbf{86.9}_{\hspace{0.05cm}1.9}$ \\
\addlinespace
\midrule
\multicolumn{5}{l}{\hspace{-0.22cm} \mycfs{10} \emph{Fact Retrieval}} \\
\multirow{2}{*}{LAMA} & $2.7$B & $14.0_{\hspace{0.05cm}0.0}$ & $\textbf{22.7}_{\hspace{0.05cm}0.0}$ & $29.7_{\hspace{0.05cm}1.8}$ & $\textbf{31.6}_{\hspace{0.05cm}1.3}$ & $35.8_{\hspace{0.05cm}3.8}$ & $\textbf{37.4}_{\hspace{0.05cm}3.4}$ & $\textbf{42.5}_{\hspace{0.05cm}1.3}$ & $\textbf{42.5}_{\hspace{0.05cm}1.4}$ \\
& $175$B & $23.5_{\hspace{0.05cm}0.0}$ & $\textbf{30.1}_{\hspace{0.05cm}0.0}$ & $48.9_{\hspace{0.05cm}2.3}$ & $\textbf{49.0}_{\hspace{0.05cm}1.4}$ & $\textbf{62.0}_{\hspace{0.05cm}2.4}$ & $61.8_{\hspace{0.05cm}2.9}$ & $\textbf{63.8}_{\hspace{0.05cm}1.0}$ & $63.6_{\hspace{0.05cm}1.3}$ \\
\midrule
\addlinespace
\multicolumn{5}{l}{\hspace{-0.22cm} \mycfs{10} \emph{Information Extraction}} \\
\multirow{2}{*}{MIT-G} & $2.7$B & $5.0_{\hspace{0.05cm}0.0}$ & $\textbf{5.7}_{\hspace{0.05cm}0.0}$ & $26.7_{\hspace{0.05cm}11.4}$ & $\textbf{37.9}_{\hspace{0.05cm}5.7}$ & $53.1_{\hspace{0.05cm}7.8}$ & $\textbf{54.7}_{\hspace{0.05cm}6.0}$ & $59.0_{\hspace{0.05cm}4.7}$ & $\textbf{59.1}_{\hspace{0.05cm}4.8}$ \\ 
\vspace{0.09cm}
& 13B & $15.0_{\hspace{0.05cm}0.0}$ & $\textbf{18.7}_{\hspace{0.05cm}0.0}$ & $47.3_{\hspace{0.05cm}3.9}$ & $\textbf{52.0}_{\hspace{0.05cm}7.9}$ & $57.9_{\hspace{0.05cm}4.8}$ & $\textbf{58.9}_{\hspace{0.05cm}4.0}$ & $59.0_{\hspace{0.05cm}4.7}$ & $\textbf{59.1}_{\hspace{0.05cm}4.8}$ \\
\multirow{2}{*}{MIT-D} & $2.7$B & $46.3_{\hspace{0.05cm}0.0}$ & $\textbf{47.0}_{\hspace{0.05cm}0.0}$ & $42.0_{\hspace{0.05cm}13.0}$ & $\textbf{53.5}_{\hspace{0.05cm}13.5}$ & $73.5_{\hspace{0.05cm}4.9}$ & $\textbf{74.1}_{\hspace{0.05cm}5.0}$ & $\textbf{75.3}_{\hspace{0.05cm}1.0}$ & $75.1_{\hspace{0.05cm}1.3}$ \\
\vspace{0.09cm}
& 13B & $36.3_{\hspace{0.05cm}0.0}$ & $\textbf{38.7}_{\hspace{0.05cm}0.0}$ & $58.6_{\hspace{0.05cm}21.4}$ & $\textbf{72.8}_{\hspace{0.05cm}4.0}$ & $75.4_{\hspace{0.05cm}1.9}$ & $\textbf{75.9}_{\hspace{0.05cm}2.1}$ & $\textbf{77.8}_{\hspace{0.05cm}0.5}$ & $\textbf{77.8}_{\hspace{0.05cm}0.5}$ \\
\multirow{2}{*}{\shortstack{ATIS-A}} & $2.7$B & $10.8_{\hspace{0.05cm}0.0}$ & $\textbf{14.0}_{\hspace{0.05cm}0.0}$ & $29.8_{\hspace{0.05cm}12.8}$ & $\textbf{33.1}_{\hspace{0.05cm}9.4}$ & $43.0_{\hspace{0.05cm}26.2}$ & $\textbf{47.3}_{\hspace{0.05cm}21.3}$ & $55.6_{\hspace{0.05cm}5.0}$ & $\textbf{58.8}_{\hspace{0.05cm}4.0}$ \\
\vspace{0.09cm}
& 13B & $49.5_{\hspace{0.05cm}0.0}$ & $\textbf{52.7}_{\hspace{0.05cm}0.0}$ & $69.6_{\hspace{0.05cm}17.4}$ & $\textbf{71.8}_{\hspace{0.05cm}17.1}$ & $67.5_{\hspace{0.05cm}10.4}$ & $\textbf{69.6}_{\hspace{0.05cm}13.4}$ & $63.4_{\hspace{0.05cm}4.6}$ & $\textbf{64.5}_{\hspace{0.05cm}4.0}$ \\
\multirow{2}{*}{ATIS-D} & $2.7$B & $6.4_{\hspace{0.05cm}0.0}$ & $\textbf{12.9}_{\hspace{0.05cm}0.0}$ & $42.3_{\hspace{0.05cm}28.8}$ & $\textbf{65.6}_{\hspace{0.05cm}20.8}$ & $75.0_{\hspace{0.05cm}6.7}$ & $\textbf{83.4}_{\hspace{0.05cm}4.2}$ & $81.0_{\hspace{0.05cm}8.8}$ & $\textbf{88.3}_{\hspace{0.05cm}3.7}$ \\
& 13B & $4.0_{\hspace{0.05cm}0.0}$ & $\textbf{5.0}_{\hspace{0.05cm}0.0}$ & $\textbf{97.9}_{\hspace{0.05cm}0.6}$ & $95.5_{\hspace{0.05cm}4.6}$ & $\textbf{98.0}_{\hspace{0.05cm}0.6}$ & $97.8_{\hspace{0.05cm}0.7}$ & $\textbf{98.8}_{\hspace{0.05cm}0.3}$ & $\textbf{98.8}_{\hspace{0.05cm}0.3}$ \\
\bottomrule
\end{tabular}
\vspace{-0.12cm}
\caption{\textbf{Contextual calibration improves accuracy across a range of tasks.} We show the mean and standard deviation across different choices of the training examples (the prompt format is fixed). The LM column indicates the GPT-3 size (see Appendix~\ref{appendix:analyzing} for GPT-2 results).
The Baseline column shows the standard approach of greedy decoding~\cite{brown2020language} and \textit{Ours} corresponds to greedy decoding after modifying the output probabilities using \methodname{}. We bold the better result of the baseline and ours. MIT-G, MIT-D, ATIS-A, and ATIS-D indicate the MIT Genre, MIT Director, ATIS Airline, and ATIS Departure Date datasets.}
\label{table:main_results}
\end{table*}

%% file: sections/60-discussion.tex
\newpage
\section{Discussion}\label{sec:discussion}

\noindent \textbf{Does Calibration Eliminate the Need to Engineer Prompts?} The motivation behind ``prompt engineering'' is that not all prompts lead to the same accuracy. Thus, one should tune the prompt's format and examples to achieve the best possible performance~\cite{brown2020language,gao2020making}. \methodnamebos{} does not eliminate the need to engineer prompts, however, it does mitigate it: \methodname{} makes the accuracy of the best, average, and worst-case prompts more similar (and higher).

\noindent \textbf{Should You Finetune in the Few-shot Setting?} We use a fixed LM with no finetuning. As mentioned in Section~\ref{sec:intro}, there are numerous reasons not to finetune: it enables rapid prototyping, provides a fully natural language interface, and is more efficient in terms of memory requirements and system complexity when serving many different tasks. Moreover, like in-context learning without contextual calibration, finetuning can be unstable in the few-shot setting~\cite{schick2020exploiting}.
Nevertheless, if these disadvantages are acceptable or avoidable, finetuning can improve accuracy over in-context learning in some cases~\cite{schick2020size,gao2020making}. An interesting direction for future work is to study the interplay between \methodname{} and finetuning, e.g., does \methodname{} alleviate the need to finetune, or vice versa?

%% file: sections/70-related.tex
\section{Related Work}

\eric{cite the GPT-3 retrieval paper}
\noindent \textbf{Few-shot Learning with Language Models} 
Recent work uses LMs to solve NLP tasks, e.g., for story cloze prediction~\cite{schwartz2017effect}, knowledge base completion~\cite{petroni2019language}, and Winograd schemas~\cite{trinh2018simple}. \citet{radford2019gpt2} and \citet{brown2020language} show that large LMs can be used to solve a myriad of tasks in a few-shot manner via in-context learning. Our paper provides a simple modification to their setting that improves performance. 
Asking LMs to complete natural language prompts is also used as a method to ``probe'' LMs, e.g., analyzing their factual~\cite{petroni2019language,jiang2019can,shin2020autoprompt} or commonsense knowledge~\cite{bosselut2019comet}. Our results suggest that these probing methods may underestimate model accuracy, and we recommend that future work take advantage of \methodname{}.\smallskip

\noindent \textbf{Volatility of Few-shot Learning in NLP} Recent work shows that when using masked language models such as BERT for zero-shot learning, the prompt format can impact accuracy~\citep{petroni2019language,jiang2019can,shin2020autoprompt}. Independent and concurrent work also shows that when finetuning masked language models on few examples, the choice of training examples can impact results~\cite{schick2020size,gao2020making}. We show that similar instabilities occur for in-context learning (i.e., no finetuning) with left-to-right language models. We also show a surprising instability associated with example ordering. Moreover, unlike past work, we analyze why these instabilities occur, and we use insights from this analysis to mitigate the issues.\smallskip

\noindent \textbf{Failures of Language Models} We identify failures when LMs are used for in-context learning (e.g., recency bias). Past work identifies similar failures when LMs are used for text generation. For example, neural LMs often repeat themselves~\cite{holtzman2019curious}, suffer from overconfidence~\cite{braverman2020calibration,jiang2020know}, suffer from recency bias~\cite{khandelwal2018lm,ravfogel2019studying}, and prefer generic responses instead of rare text~\cite{li2016diversity,logan2019barack}. Past work mitigates these degeneracies by modifying the model's output probabilities or generation schemes, e.g., explicitly preventing repetitions~\cite{paulus2017deep} or using sampling instead of greedy decoding~\cite{holtzman2019curious}.

%% file: sections/80-conclusion.tex
\section{Conclusion and Future Work}

We show that few-shot learning can be highly volatile across different choices of the prompt. Through a detailed analysis, we identify that this volatility arises from biases in LMs, e.g., their tendency to output recent or common tokens. We use these insights to develop \methodname{}---a simple procedure to adjust the model's output probabilities---which improves accuracy, reduces variance, and overall makes tools like GPT-3 more effective for end users.

Looking at the bigger picture, our results inspire two future research directions in few-shot learning for NLP. First, on the methods side, we show that good few-shot learning requires \textit{attention to detail}: small but non-trivial decisions such as calibration can greatly influence results. This makes it difficult to correctly develop and compare new methods (e.g., pretraining schemes or model architectures). We thus hope to make other few-shot learning methods more robust, and also expand our techniques to cover a wider ranger of tasks (e.g., calibration for open-ended generation).
Second, on the analysis side, our results highlight the need to understand \textit{what} GPT-3 learns from the prompt. The model has an impressive ability to improve with more training examples, however, we show that the model learns some superficial patterns such as repetition of common answers. We hope to better understand and analyze the dynamics of in-context learning in future work.

%% file: sections/acknowledgements.tex
\section*{Acknowledgements}

We thank OpenAI for providing academic access to the GPT-3 API. We thank Sewon Min, Nikhil Kandpal, Nelson Liu, Girish Sastry, Marco Tulio Ribeiro, and the members of Berkeley NLP for valuable feedback on the paper.

This work was supported by DARPA under the LwLL program/Grant No. FA8750-19-1-0504, DARPA MCS program under Contract No. N660011924033 with the United States Office Of Naval Research, DARPA and the Air Force Research Laboratory (AFRL), and NSF award \#IIS-1756023.

%% file: sections/99-appendix.tex
\clearpage
\section{Additional Results on Variance and Calibration}\label{appendix:analyzing}

Table~\ref{tab:qualitative} shows an example of the sensitivity to ordering.
\begin{table}[h]
\vspace{-0.25cm}
\centering
\mycfs{8.2}
\begin{tabular}{p{6.8cm}p{0.6cm}}
    \toprule
  {\bf Prompt} (test input not shown) & {\bf Acc.}\\
    \midrule
    Review: the whole thing 's fairly lame , making it par for the course for disney sequels .\newline
    Answer: Negative \vspace{0.1cm}\newline
    Review: this quiet , introspective and entertaining independent is worth seeking . \newline
    Answer: Positive & \multirowcell{4}{88.5\%} \\
    \midrule
    Review: this quiet , introspective and entertaining independent is worth seeking . \newline
    Answer: Positive \vspace{0.1cm}\newline
    Review: the whole thing 's fairly lame , making it par for the course for disney sequels . \newline
    Answer: Negative
    & \multirowcell{4}{51.3\%} \\
    \bottomrule
\end{tabular}
\vspace{-0.3cm}
    \caption{\underline{Top:} a prompt consisting of two training examples (the test input is not shown) that leads to good test accuracy for GPT-3 2.7B (88.5\%). \underline{Bottom:} simply \emph{reversing the order} of the two examples causes the accuracy to drop to near random chance (51.3\%).}
\label{tab:qualitative}
\end{table}

Table~\ref{table:context_words} demonstrates that the choice of content-free input
does affect accuracy, however, many good choices exist.

\begin{table}[!h]
\centering
\footnotesize
\begin{tabular}{lcc}
\toprule
\textbf{Content-free Input} & \textbf{SST-2} & \textbf{AGNews} \\
\midrule
Uncalibrated Baseline & 66.5 & 48.5 \\
\midrule
N/A & 74.2 & 64.5 \\
{[MASK]} & 74.5 & 63.8 \\
`' & 72.9 & 64.7 \\
N/A, [MASK], `' & 79.0 & 66.5 \\
the & 69.1 & 59.0 \\
abc & 77.5 & 57.3 \\
the man. & 79.4 & 62.0 \\
dasjhasjkdhjskdhds & 79.3 & 64.5 \\
nfjkhdvy84tr9bpuirvwe & 78.4 & 65.5\\
\bottomrule
\end{tabular}
\vspace{-0.19cm}
\caption{We show the accuracy for 1-shot SST-2 and 0-shot AGNews over different choices for the content-free input. The choice of content-free input matters, however, \textit{many good choices exist}. The token `' indicates the empty string. Recall that in our experiments, we ensemble over N/A, [MASK], and the empty string.}
\label{table:context_words}
\end{table}

\begin{figure*}
\captionsetup[subfigure]{labelformat=empty}
\begin{subfigure}{0.32\textwidth}
\centering
\includegraphics[trim={1.25cm 1.25cm 1.25cm 1.25cm},clip,width=1.0\textwidth]{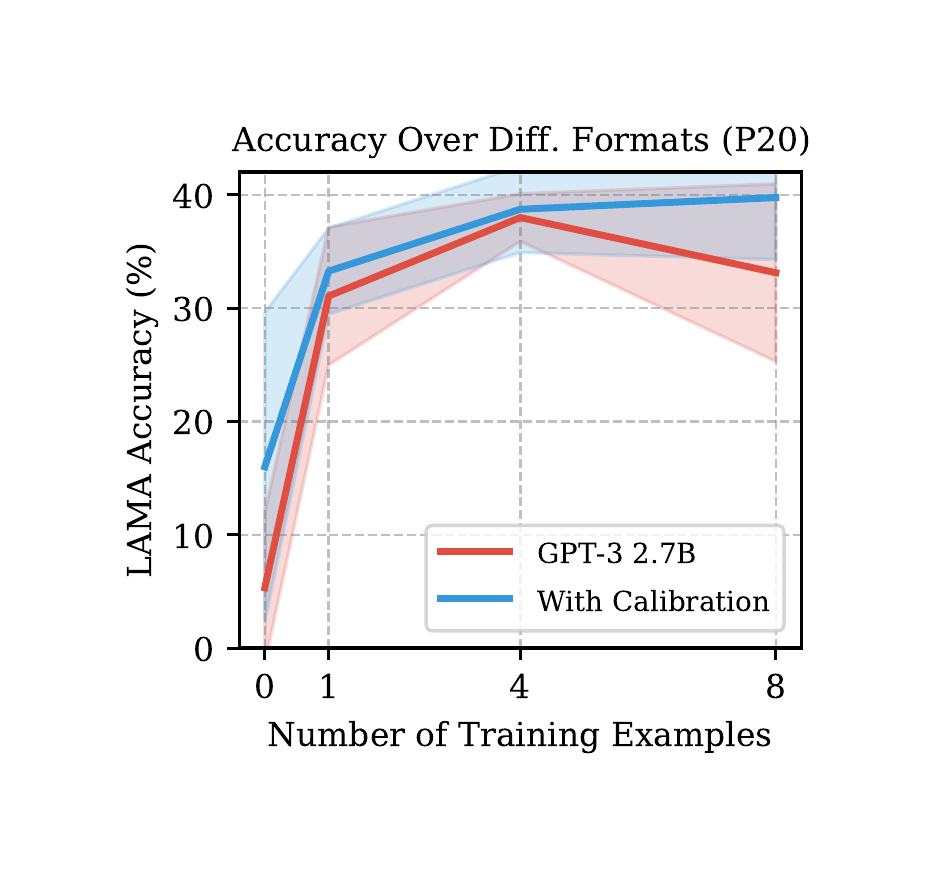}
\end{subfigure}%
\hfill
\begin{subfigure}{0.32\textwidth}
\centering
\includegraphics[trim={1.25cm 1.25cm 1.25cm 1.25cm},clip,width=1.0\textwidth]{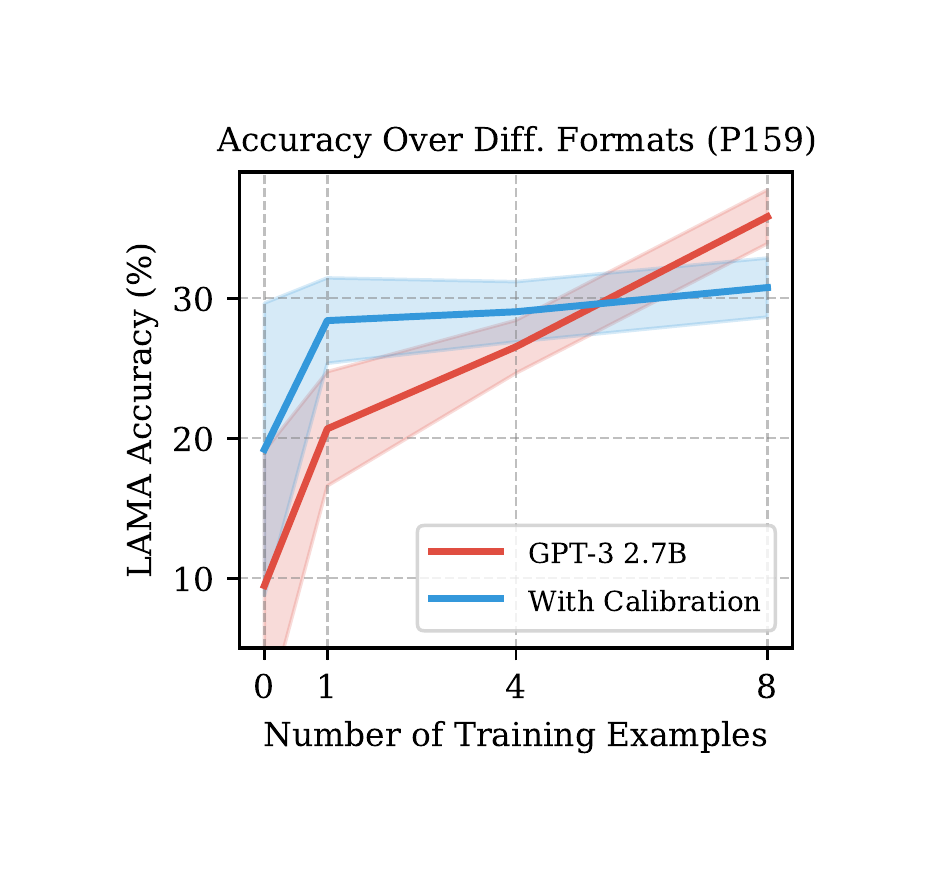}
\end{subfigure}%
\hfill
\begin{subfigure}{0.32\textwidth}
\centering
\includegraphics[trim={1.25cm 1.25cm 1.25cm 1.25cm},clip,width=1.0\textwidth]{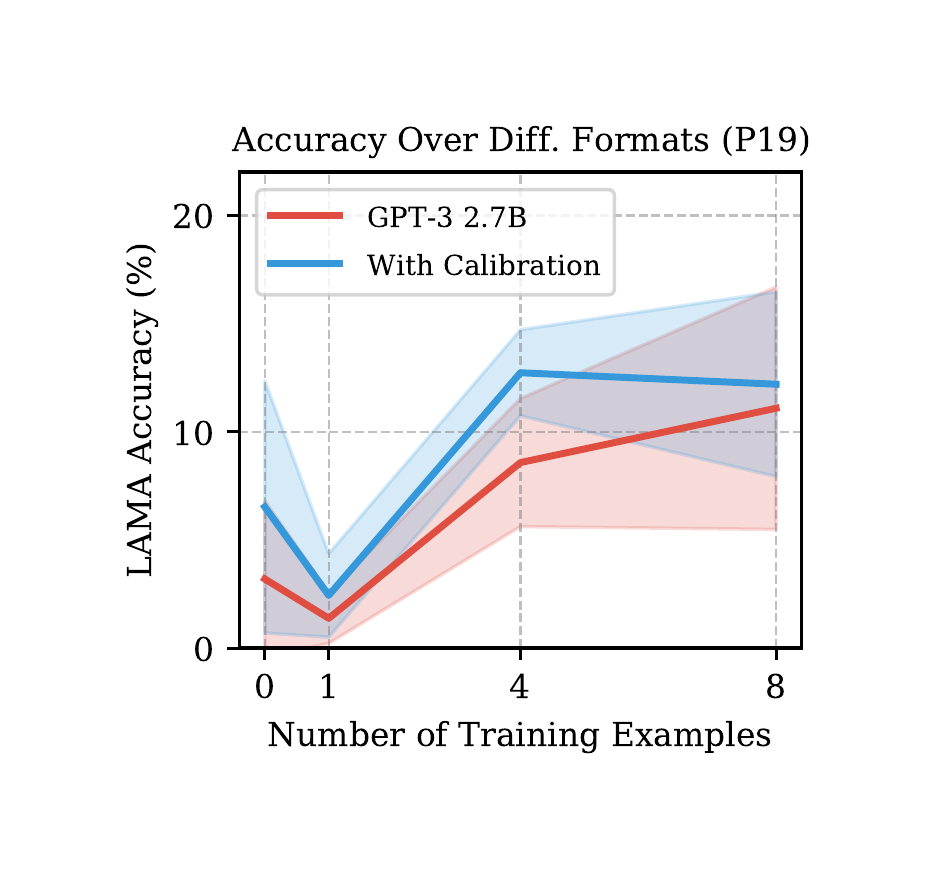}
\end{subfigure}%
\vspace{-0.2cm}
\caption{\methodnamebos{} improves GPT-3's accuracy across various prompt formats for LAMA. We plot GPT-2 2.7B's mean accuracy over 15 different formats for the LAMA ``place of death'' relation (P20), ``Headquarter Location'' relation (P159), and ``place of birth'' relation (P19).}
\label{fig:format_appendix}
\end{figure*}

Figure~\ref{fig:format_appendix} shows how GPT-3 accuracy changes as the prompt format is varied for LAMA, with and without calibration.
 
Table~\ref{table:main_results_gpt2} shows the effect of calibration for GPT-2.

\input{sections/991-gpt2-table.tex}

\section{Prompt Formats Used}\label{appendix:format}

Tables~\ref{tab:format1} and \ref{tab:format2} show the default prompt format used for all tasks. Table~\ref{tab:sst2_format_exploration} shows the 15 different formats used when studying the effect of prompt format for SST-2.

\begin{table*}
\centering
\mycfs{8.2}
\begin{tabular}{p{1.0cm}p{10.2cm}p{4.6cm}}
\toprule
\textbf{Task} & \textbf{Prompt} & \textbf{Label Names} \\
\midrule
{SST-2} & Review: This movie is amazing! \newline Sentiment: Positive
\vspace{3.9pt}\hspace{-0.1cm}

Review: Horrific movie, don't see it. \newline Sentiment: &  Positive, Negative \\

\midrule
AGNews & Article: USATODAY.com - Retail sales bounced back a bit in July, and new claims for jobless benefits fell last week, the government said Thursday, indicating the economy is improving from a midsummer slump. \newline Answer: Business
\vspace{3.9pt}\hspace{-0.1cm}

Article: New hard-drive based devices feature color screens, support for WMP 10. \newline Answer: &  World, Sports, Business, Technology \\

\midrule
TREC & Classify the questions based on whether their answer type is a Number, Location, Person, Description, Entity, or Abbreviation. \vspace{3.9pt}\hspace{-0.1cm}

Question: How did serfdom develop in and then leave Russia? \newline
Answer Type: Description
\vspace{3.9pt}\hspace{-0.1cm}

Question: When was Ozzy Osbourne born? \newline
Answer Type: &  Number, Location, Person, Description, Entity, Abbreviation \\
\midrule
DBPedia & Classify the documents based on whether they are about a Company, School, Artist, Athlete, Politician, Transportation, Building, Nature, Village, Animal, Plant, Album, Film, or Book.
\vspace{3.9pt}\hspace{-0.1cm}

Article: Geoffrey D. Falksen (born July 31 1982) is an American steampunk writer. \newline
Answer: Artist
\vspace{3.9pt}\hspace{-0.1cm}

Article: The Perrin River is a 1.3-mile-long (2.1 km) tidal river in the U.S. state of Virginia. It is a small inlet on the north shore of the York River near that river's mouth at Chesapeake Bay. \newline
Answer: &  Company, School, Artist, Athlete, Politician, Transportation, Building, Nature, Village, Animal, Plant, Album, Film, Book \\

\midrule
CB & But he ended up eating it himself. I was reluctant to kiss my mother, afraid that somehow her weakness and unhappiness would infect me. Naturally I didn't think for a minute that my life and spirit could stimulate her.\newline
question: her life and spirit could stimulate her mother. True, False, or Neither?\newline
answer: Neither
\vspace{3.9pt}\hspace{-0.1cm}

Valence the void-brain, Valence the virtuous valet. Why couldn't the figger choose his own portion of titanic anatomy to shaft? Did he think he was helping?\newline
question: Valence was helping. True, False, or Neither?\newline
answer: & True, False, Neither \\

\midrule
RTE & Others argue that Mr. Sharon should have negotiated the Gaza pullout - both to obtain at least some written promises of better Palestinian behavior, and to provide Mr. Abbas with a prime prize to show his people that diplomacy, not violence, delivered Gaza.\newline
question: Mr. Abbas is a member of the Palestinian family. True or False?\newline
answer: False
\vspace{3.9pt}\hspace{-0.1cm}

The program will include Falla's "Night in the Gardens of Spain," Ravel's Piano Concerto in G, Berlioz's Overture to "Beatrice and Benedict," and Roy Harris' Symphony No. 3.\newline
question: Beatrice and Benedict is an overture by Berlioz. True or False?\newline
answer: & True, False \\
\bottomrule
\end{tabular}
\vspace{-0.2cm}
\caption{The prompts used for text classification. We show one training example per task for illustration purposes. The right column shows the label names (to make predictions, we check the LM's probability for these tokens).}
\label{tab:format1}
\end{table*}

\begin{table*}
\centering
\mycfs{8.2}
\begin{tabular}{p{2.0cm}p{13.0cm}}
\toprule
\textbf{Task} & \textbf{Prompt}\\
\midrule
LAMA & Alexander Berntsson was born in Sweden \vspace{3.9pt}\hspace{-0.1cm}

Khalid Karami was born in \\

\midrule
ATIS \newline (Airline) & Sentence: what are the two american airlines flights that leave from dallas to san francisco in the evening
\newline Airline name: american airlines \vspace{3.9pt}\hspace{-0.1cm}

Sentence: list a flight on american airlines from toronto to san diego \newline Airline name: \\
\midrule
ATIS \newline (Depart Date) & Sentence: please list any flight available leaving oakland california tuesday arriving philadelphia wednesday
\newline Depart date - Day name: tuesday \vspace{3.9pt}\hspace{-0.1cm}

Sentence: show me all all flights from pittsburgh to atlanta on wednesday which leave before noon and serve breakfast \newline Depart date - Day name: \\
\midrule
MIT Movies \newline (Genre) & Sentence: last to a famous series of animated movies about a big green ogre and his donkey and cat friends
\newline Genre: animated
\vspace{3.9pt}\hspace{-0.1cm}

Sentence: what is a great comedy featuring the talents of steve carell as a loser looking for a friend \newline Genre: \\

\midrule
MIT Movies \newline (Director) & Sentence: in 2005 director christopher nolan rebooted a legendary dc comics superhero with a darker grittier edge in which movie
\newline Director: christopher nolan
\vspace{3.9pt}\hspace{-0.1cm}

Sentence: what 1967 mike nichols film features dustin hoffman in romantic interludes with anne bancroft as mrs robinson \newline Director: \\
\bottomrule
\end{tabular}
\vspace{-0.2cm}
\caption{The prompts used for generation tasks. We show one training example per task for illustration purposes.}
\label{tab:format2}
\end{table*}

\begin{table*}
\centering
\mycfs{8.2}
\aboverulesep = 0.205mm
\belowrulesep = 0.405mm
\setlength{\tabcolsep}{2pt}
\begin{tabular}{cp{12.7cm}p{2.3cm}}
\toprule
\textbf{Format ID} & \textbf{Prompt} & \textbf{Label Names} \\
\midrule
1 & Review: This movie is amazing! \newline\vspace{2.6pt}\hspace{-0.1cm} Answer: Positive \newline Review: Horrific movie, don't see it. \newline Answer: &  Positive, Negative \\

\midrule
2 & Review: This movie is amazing! \newline\vspace{2.6pt}\hspace{-0.1cm} Answer: good \newline Review: Horrific movie, don't see it. \newline Answer: &  good, bad \\

\midrule
3 & My review for last night's film: This movie is amazing! The critics agreed that this movie was good \vspace{2.6pt} \newline My review for last night's film: Horrific movie, don't see it. The critics agreed that this movie was & good, bad \\

\midrule
4 & Here is what our critics think for this month's films. \vspace{2.6pt}\newline \vspace{2.6pt}\hspace{-0.1cm} 
One of our critics wrote "This movie is amazing!". Her sentiment towards the film was positive. \newline 
One of our critics wrote "Horrific movie, don't see it". Her sentiment towards the film was & positive, negative \\

\midrule
5 & Critical reception [ edit ] \vspace{2.6pt}\newline
In a contemporary review, Roger Ebert wrote "This movie is amazing!". Entertainment Weekly agreed, and the overall critical reception of the film was good.\vspace{2.6pt}\newline
In a contemporary review, Roger Ebert wrote "Horrific movie, don't see it". Entertainment Weekly agreed, and the overall critical reception of the film was & good, bad \\

\midrule
6 & Review: This movie is amazing! \newline\vspace{2.6pt}\hspace{-0.1cm} Positive Review? Yes \newline Review: Horrific movie, don't see it. \newline Positive Review? & Yes, No \\

\midrule
7 & Review: This movie is amazing! \newline
Question: Is the sentiment of the above review Positive or Negative? \newline\vspace{2.6pt}\hspace{-0.1cm}
Answer: Positive \newline Review: This movie is amazing! \newline
Question: Is the sentiment of the above review Positive or Negative? \newline
Answer: & Positive, Negative \\

\midrule
8 & Review: This movie is amazing! \newline
Question: Did the author think that the movie was good or bad? \newline\vspace{2.6pt}\hspace{-0.1cm}
Answer: good \newline Review: This movie is amazing! \newline
Question: Did the author think that the movie was good or bad? \newline
Answer: & good, bad \\

\midrule
9  & Question: Did the author of the following tweet think that the movie was good or bad? \newline
Tweet: This movie is amazing! \newline\vspace{2.6pt}\hspace{-0.1cm}
Answer: good \newline Question: Did the author of the following tweet think that the movie was good or bad?  \newline
Tweet: Horrific movie, don't see it \newline
Answer: & good, bad \\

\midrule
10 & This movie is amazing! My overall feeling was that the movie was good \vspace{2.6pt}\newline
Horrific movie, don't see it. My overall feeling was that the movie was & good, bad \\

\midrule
11  & This movie is amazing! I liked the movie. \vspace{2.6pt}\newline  Horrific movie, don't see it. I & liked, hated \\

\midrule
12 & This movie is amazing! My friend asked me if I would give the movie 0 or 5 stars, I said 5 \vspace{2.6pt} \newline Horrific movie, don't see it. My friend asked me if I would give the movie 0 or 5 stars, I said & 0, 5 \\

\midrule
13 & Input: This movie is amazing! \newline\vspace{2.6pt}\hspace{-0.1cm} Sentiment: Positive \newline Input: Horrific movie, don't see it.
\newline Sentiment: & Positive, Negative \\

\midrule
14 & Review: This movie is amazing! \newline\vspace{2.6pt}\hspace{-0.1cm} Positive: True \newline Review: Horrific movie, don't see it. \newline Positive: &  True, False \\

\midrule
15 & Review: This movie is amazing! \newline\vspace{2.6pt}\hspace{-0.1cm} Stars: 5 \newline Review: Horrific movie, don't see it. \newline Stars: &  5, 0 \\
\bottomrule
\end{tabular}
\vspace{-0.2cm}
\caption{The different prompt formats used when studying the effect of format for SST-2. We show one training example for illustration.}
\label{tab:sst2_format_exploration}
\aboverulesep = 0.605mm
\belowrulesep = 0.984mm
\end{table*}

%% file: sections/991-gpt2-table.tex
\begin{table*}[b]
\centering
\mycfs{10}
\setlength{\tabcolsep}{5pt}
\begin{tabular}{lcllllllll}
\toprule
\textbf{Dataset} & \textbf{LM} &  \multicolumn{2}{c}{\textbf{0-shot}} &       \multicolumn{2}{c}{\textbf{1-shot}} & \multicolumn{2}{c}{\textbf{4-shot}} & \multicolumn{2}{c}{\textbf{8-shot}} \\
\cmidrule(lr){3-4}
\cmidrule(lr){5-6}
\cmidrule(lr){7-8}
\cmidrule(lr){9-10}
& & Baseline & {~~Ours} & Baseline & {~~~Ours} & Baseline & {~~~Ours} & Baseline & {~~~Ours} \\
\midrule
\multicolumn{5}{l}{\hspace{-0.22cm} \mycfs{10} \emph{Text Classification}} \\
\vspace{0.09cm}
\multirow{1}{*}{AGNews} & GPT-2 & $44.0_{\hspace{0.05cm}\text{0.0}}$ & $\textbf{60.0}_{\hspace{0.05cm}\text{0.0}}$ & $45.4_{\hspace{0.05cm}8.4}$ & $\textbf{67.9}_{\hspace{0.05cm}5.7}$ & $44.6_{\hspace{0.05cm}12.2}$ & $\textbf{58.0}_{\hspace{0.05cm}13.6}$ & $57.1_{\hspace{0.05cm}11.6}$ & $\textbf{63.1}_{\hspace{0.05cm}7.3}$ \\
\vspace{0.09cm}
\multirow{1}{*}{TREC} & GPT-2 & $24.0_{\hspace{0.05cm}\text{0.0}}$ & $\textbf{37.3}_{\hspace{0.05cm}\text{0.0}}$ & $21.5_{\hspace{0.05cm}5.2}$ & $\textbf{41.1}_{\hspace{0.05cm}2.6}$ & $23.1_{\hspace{0.05cm}5.9}$ & $\textbf{44.2}_{\hspace{0.05cm}2.2}$ & $32.7_{\hspace{0.05cm}7.5}$ & $\textbf{44.1}_{\hspace{0.05cm}3.6}$ \\
\vspace{0.09cm}
\multirow{1}{*}{CB} & GPT-2 & $\textbf{44.6}_{\hspace{0.05cm}\text{0.0}}$ & $17.9_{\hspace{0.05cm}\text{0.0}}$ & $\textbf{49.6}_{\hspace{0.05cm}10.0}$ & $47.1_{\hspace{0.05cm}12.2}$ & $40.0_{\hspace{0.05cm}8.3}$ & $\textbf{55.4}_{\hspace{0.05cm}7.3}$ & $48.9_{\hspace{0.05cm}5.7}$ & $\textbf{63.2}_{\hspace{0.05cm}1.4}$ \\
\vspace{0.09cm}
\multirow{1}{*}{RTE} & GPT-2 & $\textbf{51.0}_{\hspace{0.05cm}0.0}$ & $48.5_{\hspace{0.05cm}0.0}$ & $\textbf{57.6}_{\hspace{0.05cm}2.1}$ & $56.3_{\hspace{0.05cm}2.4}$ & $53.2_{\hspace{0.05cm}6.0}$ & $\textbf{57.5}_{\hspace{0.05cm}1.8}$ & $54.9_{\hspace{0.05cm}3.0}$ & $\textbf{57.7}_{\hspace{0.05cm}1.29}$ \\
\vspace{0.09cm}
\multirow{1}{*}{SST-2} & GPT-2 & $60.0_{\hspace{0.05cm}\text{0.0}}$ & $\textbf{82.0}_{\hspace{0.05cm}\text{0.0}}$ & $66.7_{\hspace{0.05cm}17.9}$ & $\textbf{73.0}_{\hspace{0.05cm}11.4}$ & $64.9_{\hspace{0.05cm}8.4}$ & $\textbf{73.8}_{\hspace{0.05cm}10.9}$ & $54.5_{\hspace{0.05cm}4.6}$ & $\textbf{64.6}_{\hspace{0.05cm}8.8}$ \\
\multirow{1}{*}{DBPedia} & GPT-2 & $\textbf{64.3}_{\hspace{0.05cm}\text{0.0}}$ & $58.3_{\hspace{0.05cm}\text{0.0}}$ & $33.6_{\hspace{0.05cm}18.9}$ & $\textbf{69.5}_{\hspace{0.05cm}9.4}$ & $53.0_{\hspace{0.05cm}14.8}$ & $\textbf{75.3}_{\hspace{0.05cm}8.1}$ & $66.0_{\hspace{0.05cm}3.6}$ & $\textbf{74.3}_{\hspace{0.05cm}8.7}$ \\
\addlinespace
\midrule
\multicolumn{5}{l}{\hspace{-0.22cm} \mycfs{10} \emph{Fact Retrieval}} \\
\multirow{1}{*}{LAMA} & GPT-2 & $14.0_{\hspace{0.05cm}\text{0.0}}$ & $\textbf{22.7}_{\hspace{0.05cm}\text{0.0}}$ & $29.7_{\hspace{0.05cm}1.8}$ & $\textbf{31.6}_{\hspace{0.05cm}1.3}$ & $35.8_{\hspace{0.05cm}3.8}$ & $\textbf{37.4}_{\hspace{0.05cm}3.4}$ & $\textbf{42.5}_{\hspace{0.05cm}1.3}$ & $\textbf{42.5}_{\hspace{0.05cm}1.4}$ \\
\midrule
\addlinespace
\multicolumn{5}{l}{\hspace{-0.22cm} \mycfs{10} \emph{Information Extraction}} \\
\vspace{0.09cm}
\multirow{1}{*}{MIT-G} & GPT-2 & $7.7_{\hspace{0.05cm}\text{0.0}}$ & $\textbf{10.0}_{\hspace{0.05cm}\text{0.0}}$ & $32.9_{\hspace{0.05cm}10.0}$ & $\textbf{41.2}_{\hspace{0.05cm}4.1}$ & $44.3_{\hspace{0.05cm}6.5}$ & $\textbf{47.7}_{\hspace{0.05cm}5.8}$ & $56.9_{\hspace{0.05cm}2.5}$ & $\textbf{59.5}_{\hspace{0.05cm}2.5}$ \\
\vspace{0.09cm}
\multirow{1}{*}{MIT-D} & GPT-2 & $29.3_{\hspace{0.05cm}\text{0.0}}$ & $\textbf{41.7}_{\hspace{0.05cm}\text{0.0}}$ & $26.2_{\hspace{0.05cm}10.5}$ & $\textbf{58.8}_{\hspace{0.05cm}4.8}$ & $70.5_{\hspace{0.05cm}2.5}$ & $\textbf{75.4}_{\hspace{0.05cm}1.8}$ & $77.1_{\hspace{0.05cm}4.4}$ & $\textbf{78.1}_{\hspace{0.05cm}3.9}$ \\
\vspace{0.09cm}
\multirow{1}{*}{\shortstack{ATIS-A}} & GPT-2 & $15.1_{\hspace{0.05cm}\text{0.0}}$ & $\textbf{35.5}_{\hspace{0.05cm}\text{0.0}}$ & $41.5_{\hspace{0.05cm}11.7}$ & $\textbf{51.4}_{\hspace{0.05cm}7.5}$ & $55.1_{\hspace{0.05cm}18.9}$ & $\textbf{65.8}_{\hspace{0.05cm}11.7}$ & $63.4_{\hspace{0.05cm}10.6}$ & $\textbf{69.9}_{\hspace{0.05cm}10.4}$ \\
\multirow{1}{*}{ATIS-D} & GPT-2 & $1.0_{\hspace{0.05cm}\text{0.0}}$ & $\textbf{2.5}_{\hspace{0.05cm}\text{0.0}}$ & $62.3_{\hspace{0.05cm}9.2}$ & $\textbf{68.7}_{\hspace{0.05cm}4.3}$ & $81.1_{\hspace{0.05cm}3.6}$ & $\textbf{83.2}_{\hspace{0.05cm}7.2}$ & $81.8_{\hspace{0.05cm}4.5}$ & $\textbf{83.9}_{\hspace{0.05cm}5.0}$ \\
\bottomrule
\end{tabular}
\vspace{-0.12cm}
\caption{\textbf{Contextual calibration improves accuracy for GPT-2.} This table is analogous to Table~\ref{table:main_results} but shows results for GPT-2 XL.}
\label{table:main_results_gpt2}
\end{table*}